\documentclass[sigconf,nonacm]{acmart}

\usepackage{microtype}
\usepackage{graphicx}
\usepackage{subcaption}
\usepackage{booktabs}
\usepackage{multirow}
\usepackage{enumitem}
\usepackage{xcolor}

\setcopyright{none}
\copyrightyear{2027}
\acmYear{2027}
\acmConference[]{}
\acmDOI{}
\acmISBN{}
\begin{document}
\title{Embodied Graph: A Relational World Model for Long-Horizon Tasks}

\renewcommand{\shortauthors}{Wu et al.}

\makeatletter
\def\@mkauthors{%
  \gdef\@currentauthors{}%
  \global\setbox\mktitle@bx=\vbox{%
    \noindent\unvbox\mktitle@bx\par\medskip
    \centering
    {\Large
      Zongyuan Wu, Xin Wang\footnotemark[1], Tongtong Feng,
      Yuan Meng, Zeyang Zhang,
      Ling Feng,
      Wenwu Zhu\footnotemark[1]\par
    }
    \vspace{0.5em}
    {\normalsize Tsinghua University\par}
    \bigskip
  }%
}
\authornote{Corresponding authors: Xin Wang
(xin\_wang@tsinghua.edu.cn) and Wenwu Zhu
(wwzhu@tsinghua.edu.cn).}
\makeatother

\begin{abstract}
Long-horizon tasks in embodied and web environments involve extended decision sequences and delayed feedback, making accurate planning crucial for success. World models tackle this by learning latent dynamics for imagination-based planning. However, existing approaches merely rely on \textit{dense} pixel-level observations, which forces the model to absorb redundant information while overlooking relations among task-relevant entities. Consequently, small errors may accumulate over long imagined rollouts and gradually corrupt the predictions. In this paper, we explore world models that imagine over sparse state abstractions, which poses two key challenges: 1) how to remove redundant information and form a compact state abstraction; 2) how to incorporate the relational structure to learn reliable long-horizon dynamics. To address these challenges, we propose \textbf{Embodied Graph}, a relational state abstraction that supports structured dynamics modeling and long-horizon imagination, and the \underline{\textbf{R}}elational \underline{\textbf{W}}orld \underline{\textbf{M}}odel (\textbf{RWM}), which performs imagination directly over graph states. RWM consists of (i) a \emph{Task-Aware Embodied Graph Constructor} that filters redundant details and maintains graph memory through online updates, and (ii) an \emph{Embodied Graph Recurrent State Space Model} that learns node-wise latent dynamics while capturing inter-node dependencies. The imagined graph states provide predictions of action consequences to guide decision-making. Experiments on Minecraft and Crafter demonstrate the effectiveness of RWM in embodied environments. Notably, in Minecraft, RWM is the only method that obtains diamonds within the given budget. Experiments on web interaction further illustrate the applicability of relational state prediction to the decision making process of web agents.
\end{abstract}



\maketitle

\section{Introduction}
Embodied long-horizon tasks~\cite{feng2025u2udatascalableswarmuavs,feng2025evoagentselfevolvingagentcontinual,guo2024,Shen2025DETACHCL} are interactive tasks in which agents must maintain coherent planning and decision-making over an extended interaction horizon to achieve the final goal. Despite recent progress, these tasks remain challenging in open dynamic environments~\cite{feng2026selfevolvingembodiedai}, ranging from manipulation~\cite{ma2024hierarchical,yang2025lohovla}, navigation~\cite{liu2024ver,hong2025gsa} to task planning~\cite{wang-etal-2025-world,fei2025embodiedplannerr1}, largely due to sparse rewards and delayed credit assignment over long time spans~\cite{pignatelli2023creditassignment}. World models have emerged as a promising approach for such tasks by learning internal representations and predicting future states of the environment for imagination-based planning and control~\citep{shang2026worldarenaunifiedbenchmarkevaluating,Li_2025_CVPR, hafner2025dreamerv3, ren2025surfer, feng2025embodiedaillmsworld}.

\begin{figure}[t]
\vspace{20pt}
\begin{center}
\includegraphics[width=\linewidth]{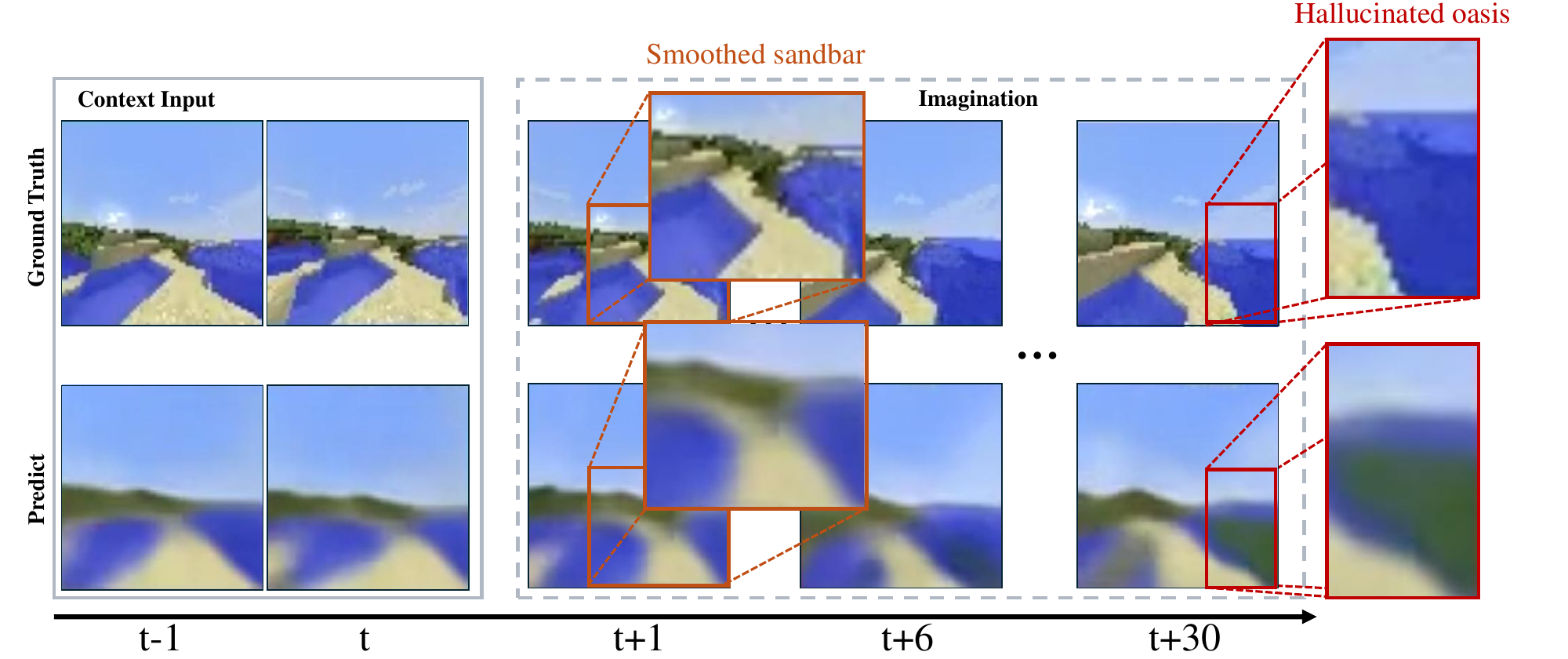}
\end{center}
\caption{Imagination drift in a pixel-level world model. A minor early inaccuracy (a smoothed sandbar) can accumulate into a severe deviation (a hallucinated oasis) that would lead the agent toward an implausible state.}
\label{fig:drift}
\end{figure}

Nevertheless, existing world models are typically trained with \textit{dense} supervision from high-dimensional observations, which injects massive redundant details \cite{kim2025make,deng2021dreamerproreconstructionfreemodelbasedreinforcement, burchi2024mudreamerlearningpredictiveworld} and obscures the spatio-temporal relations that matter for long-horizon prediction and planning \cite{kipf2020contrastivelearningstructuredworld}. The amount of visual change does not necessarily reflect its importance for decision-making. A visual change can alter many pixels while leaving the underlying object states unchanged, whereas acquiring a needed item may produce only a small visual difference. Dense prediction objectives therefore devote capacity to appearance changes while overlooking state transitions that determine subsequent actions, which can lead to errors in predicting task-relevant state changes. Small errors may then accumulate over long rollouts~\cite{talvitie2017selfcorrectingmodelsmodelbasedreinforcement}, forming imagination drift to corrupt imagined trajectories and shortening the effective imagination horizon. In Fig.~\ref{fig:drift}, a smoothed sandbar compounds into a hallucinated oasis, and a policy trained on such rollouts is steered toward states that the real environment never offers.

For this reason, we seek to use sparse relational graphs as state abstractions, alleviating error accumulation through a  compact state representation. Unlike dense representations, graphs offer a structured and compact abstraction of objects and their relations, which can naturally filter out irrelevant details and focus on critical interactions. For instance, on interactive webpages, such an abstraction can capture interactive page elements and their relations, supporting predictions of changes in field values and control states that inform subsequent action choices. However, leveraging graph representations for long-horizon tasks introduces several challenges: (1) since environments are open and dynamic, objects and relations may appear, disappear, or alter their attributes, it is challenging to distill informative structure and update the graph online while maintaining consistency and stability. (2) It is also challenging to learn stable graph dynamics whose rollouts incur less compounding error, thereby extending the effective imagination horizon, i.e., how far the model can be rolled out before its error grows substantially.

To tackle these challenges, we introduce \textbf{Embodied Graph}, a sparse relational state abstraction for open dynamic environments. In Embodied Graph, nodes represent the agent and environment entities, edges encode the relations along which the effects of interactions propagate, and an interaction is an agent action applied to a target node. On top of it, we build the \underline{\textbf{R}}elational \underline{\textbf{W}}orld \underline{\textbf{M}}odel (\textbf{RWM}), a \emph{graph world model} with two learned components: (i) a \emph{Task-Aware Embodied Graph Constructor (TA-EGC)} that maintains the embodied graph as a bounded global memory updated from online observations, and (ii) an \emph{Embodied Graph Recurrent State Space Model (EG-RSSM)} that learns node-wise latent dynamics with relational propagation and imagines over graph states. Experiments in Embodied environments (Minecraft \citep{hill2023mineplanner} and Crafter \citep{hafner2021crafter}) show that RWM outperforms state-of-the-art baselines, with the largest gains on long-horizon targets, and its open-loop prediction error grows much more slowly than that of DreamerV3. And further experiments on web agents~\cite{zhou2024webarena} illustrate the effectiveness of RWM on web agents. Our contributions are:

\begin{itemize}
    \item We propose \textbf{Embodied Graph}, a graph-based paradigm for modeling environment dynamics in long-horizon tasks.
    \item We propose \textbf{Relational World Model} (\textbf{RWM}), a graph-based world model framework consists of a \emph{Task-Aware Embodied Graph Constructor} for maintaining a sparse global embodied graph, and an \emph{Embodied Graph Recurrent State Space Model} that performs node-wise latent dynamics learning while capturing inter-node dependencies. 
    \item Through extensive experiments, we demonstrate that the proposed \textbf{RWM} achieves superior performance over strong baselines in challenging long-horizon tasks on both of the embodied control and web agents.
\end{itemize}

\section{Embodied Graph}
\label{sec:method:eg}
As structured abstractions of complex relational patterns, graphs are well-suited for modeling the intricate relationships between agents and entities in their environments. However, traditional graph approaches such as 3D scene graphs cannot adequately model dynamic interactions. To address this limitation, we introduce the \textbf{Embodied Graph} (\textbf{EG}), an interactive graph formalism designed to capture environmental relational dynamics. We formally define the embodied graph and the associated task formulation as follows.

\textbf{Embodied Graph.} We denote the embodied graph at time $t$ as $G^t(\mathcal{V}^t, \mathcal{E}^t, \mathcal{I}^t)$, where $\mathcal{V}^t$, $\mathcal{E}^t$, and $\mathcal{I}^t$ represent the sets of nodes, edges, and interactions, respectively. The complete embodied graph is expressed as $\mathcal{G}(\mathcal{V}, \mathcal{E}, \mathcal{I}) = \{G^t(\mathcal{V}^t, \mathcal{E}^t, \mathcal{I}^t)\}_{t=1}^T$, where $\mathcal{V} = \bigcup_{t=1}^T \mathcal{V}^t$, $\mathcal{E} = \bigcup_{t=1}^T \mathcal{E}^t$, and $\mathcal{I} = \bigcup_{t=1}^T \mathcal{I}^t$.

\textbf{Node, Edge, and Interaction Sets.} Nodes in the embodied graph encompass both agents and environment entities, where agents are capable of actively interacting with entities. This heterogeneous node structure is represented as $\mathcal{V} = \{\mathcal{V}_a, \mathcal{V}_e\}$, with $\mathcal{V}_a$ and $\mathcal{V}_e$ corresponding to the agent and entity nodes, respectively. Therefore, edges are also heterogeneous and categorized into entity-entity, entity-agent, and agent-agent relationships, denoted as $\mathcal{E} = \{\mathcal{E}_{ee}, \mathcal{E}_{ea}, \mathcal{E}_{aa}\}$ (agent-agent edges exist only in multi-agent scenarios). Both nodes and edges incorporate the corresponding feature sets $\mathcal{X} = \{\mathcal{X}_a, \mathcal{X}_e\}$ and $\mathcal{S} = \{\mathcal{S}_{ee}, \mathcal{S}_{ea}, \mathcal{S}_{aa}\}$. Edges are directed, where $(u,v)\in\mathcal{E}^t$ means that the state of $u$ informs the change of $v$, a symmetric relation is represented by two opposite edges, and $\mathcal{N}(v)=\{u:(u,v)\in\mathcal{E}^t\}$ denotes the neighbors of $v$.
Unlike traditional graphs, embodied graphs involve interactions between different nodes. An interaction $A=(i,v)$ consists of an interaction type $i$ and a target node $v$. Due to the heterogeneity of the nodes, similar to the edges, the interactions can be divided as $\mathcal{I} = \{\mathcal{I}_{ee}, \mathcal{I}_{ea}, \mathcal{I}_{aa}\}$. The corresponding interaction features are denoted as $\mathcal{R} = \{\mathcal{R}_{ee}, \mathcal{R}_{ea}, \mathcal{R}_{aa}\}$.

\textbf{Prediction Tasks.} We consider a partially observable environment in which the agent receives observations $o_t$ and aims to maximize its expected return. The embodied graph is maintained from the observations, $G^t=C_\phi(G^{t-1},o_t)$, and provides a structured state abstraction to predict environment evolution under interactions:
\begin{equation}
p_\Theta\big(G^{t+1}\mid G^{\le t},A^{\le t}\big),\qquad A^t=(i^t,v_{tar}^t)\in\mathcal{I}^t,
\label{eq:gwm}
\end{equation}
where $p_\Theta$ is the learnable dynamics model and $A^t$ is the chosen interaction with its target node. The model is learned by maximizing the likelihood of the observed next graphs.
\begin{equation}
\min_\Theta \ \mathcal{L}(\Theta)=-\sum_t\log p_\Theta\big(G^{t+1}\mid G^{\le t},A^{\le t}\big),
\end{equation}

\section{Relational World Model}
\label{sec:method}
This section presents the \textbf{Relational World Model (RWM)}. RWM consists of the Task-Aware Embodied Graph Constructor (Section~\ref{sec:method:degc}), which implements $C_\phi$, and the Embodied Graph Recurrent State Space Model (Section~\ref{sec:method:egrssm}), which implements $p_\Theta$ and performs imagination. Section~\ref{sec:decision} describes how RWM is used for decision making in embodied control and web agents.

\begin{figure*}[h]
\begin{center}
\includegraphics[width=\linewidth]{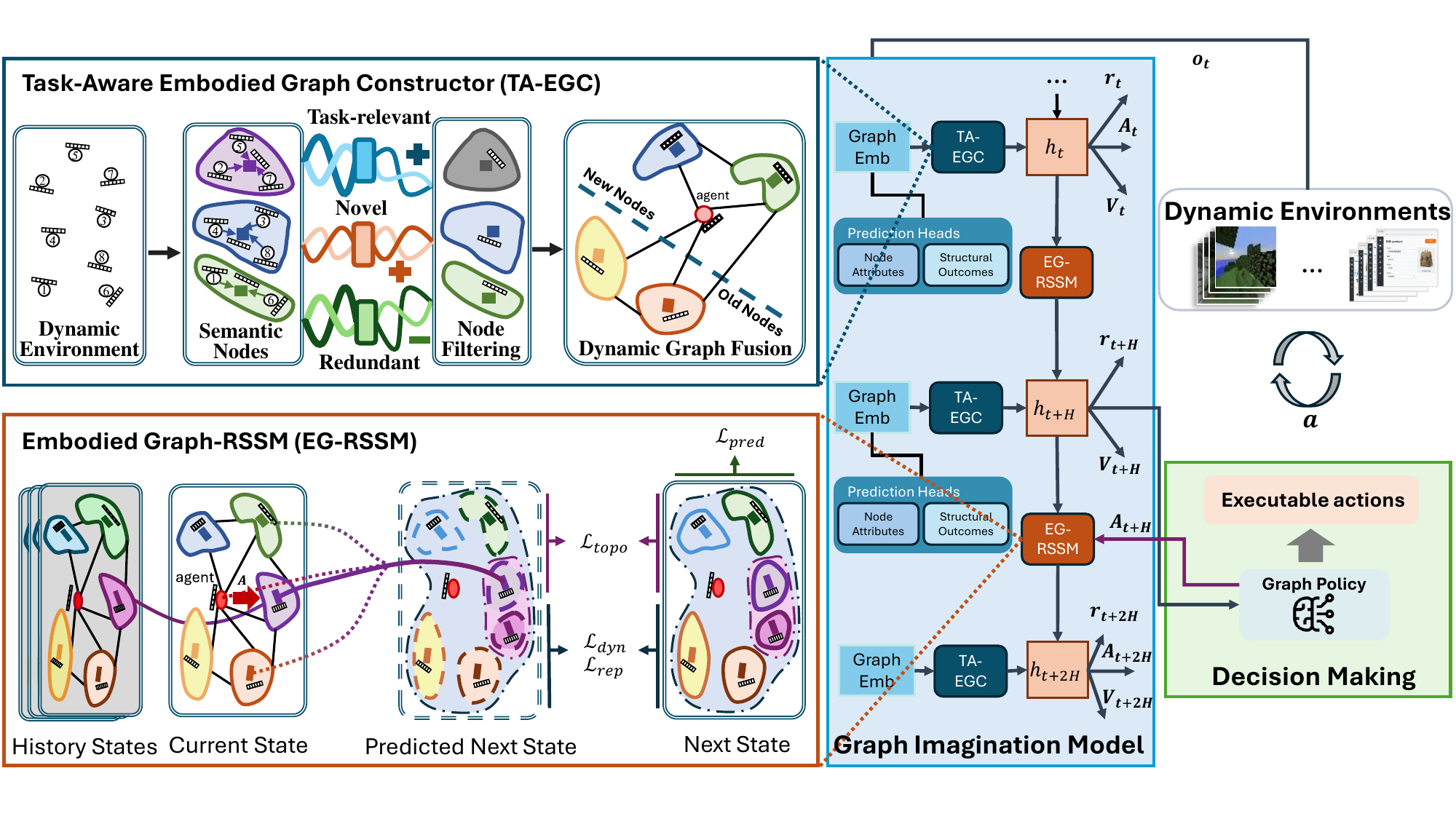}
\end{center}
\caption{Overall framework of RWM. A Task-Aware Embodied Graph Constructor builds and continuously updates a compact embodied graph, while an Embodied Graph Recurrent State Space Model learns action-conditioned, node-wise dynamics through relational propagation. The graph imagination model predicts future node attributes and structural outcomes, providing forecasts of action consequences to guide decision-making in dynamic environments.}
\label{fig:framework}
\end{figure*}

\subsection{Task-Aware Embodied Graph Constructor (TA-EGC)}
\label{sec:method:degc}
TA-EGC implements $C_\phi$ and maintains the embodied graph as a bounded global memory of at most $N$ nodes. At each step, a front-end converts the observation $o_t$ into candidate nodes $\hat{\mathcal{V}}^t=\{\hat{v}_i^t\}_{i=1}^{M}$ with attribute vectors $\hat{o}_i^t$. The update $(G^{t-1},\hat{\mathcal{V}}^t)\rightarrow G^t$ consists of task-aware node filtering and dynamic graph fusion.

\textbf{Task-aware node filtering.} Inserting all candidates would fill the graph with redundant fragments and transient noise. Therefore, an admission scorer rates each candidate by task relevance, novelty, and coverage, $s_i^t=\alpha_t s_{i,task}^t+\alpha_n s_{i,novel}^t+\alpha_c s_{i,cover}^t$, and the Top-$P$ candidates form the insertion set $\mathcal{U}^t$. Each score is predicted by a scorer $S_{\theta_k}$, where $k\in\mathcal{K}=\{task,novel,cover\}$, and trained with labels $y_{i,k}^t\in[0,1]$:
\begin{equation}
    \mathcal{L}_{score} = - \sum_{k \in \mathcal{K}}\lambda_k \frac{\sum_{i \in \mathcal{D}^t}w_{i,k}^t \log p_{\theta_k} (y_{i, k}^t \mid \hat{o}_i^t)}{\sum_{i \in \mathcal{D}^t}w_{i,k}^t},
\end{equation}
where $\mathcal{D}^t$ indexes the candidates at time $t$ and $\lambda_k$ are loss weights. The labels are defined in hindsight: a candidate is task-relevant if it becomes the target of an interaction within the next $W$ steps, novelty reflects how rare its semantic category is among the current candidates, and coverage favors candidates that spread insertions over the observation instead of concentrating them in one region. The weights $w_{i,k}^t$ are uniform, except that the task term is reweighted by the policy advantage $\text{Adv}_i^t$ to emphasize important steps.

\textbf{Dynamic graph fusion.} Each existing node $v_j^{t-1}$ is associated with at most one admitted candidate through an identity test. The best-matching candidate updates the node's attributes and its last-seen step $l_j$, and unmatched candidates are inserted as new nodes. When the graph reaches its capacity, the nodes unseen for the longest time are evicted. Finally, the edge set is rebuilt by a relation function, $\mathcal{E}^t=\mathcal{R}(\mathcal{V}^t)$, and the messages exchanged along the edges are learned by EG-RSSM.

\subsection{Embodied Graph Recurrent State Space Model (EG-RSSM)}
\label{sec:method:egrssm}
Most world models learn dynamics in an entangled latent state, where objects, background, and noise are mixed and inter-object relations are not modeled explicitly, which makes the learned dynamics brittle to small variations. Since our embodied graph already filters redundancy, we predict and update the environment state directly in the graph. EG-RSSM maintains a latent state for each node and learns relation-aware node-wise dynamics, propagating interaction effects along edges.

\textbf{Node-wise latent state.} 
We represent the latent state of each node $v_{j}^t$ as $\{h_j^{t}, z_j^{t}\}$, where $h_j^{t}$ is the deterministic latent state and $z_j^{t}$ is the stochastic latent state.
This node-wise factorization encourages the dynamics model $f_{dyn}$ to operate on the retained, task-relevant objects in $G^t$, avoiding entangled modeling of dense background content and thereby reducing redundancy in long-horizon prediction.

\textbf{Graph-conditioned recurrent state transition.} For each node $v_{j}^t$, conditioned on the graph-level action $A^{t-1}$ and current node attributes $o_j^t$, we obtain node embeddings via $x_j^{t}=f_{enc}(o_j^{t})$, and further incorporate relational information by computing a context representation $c_j^{t}$ through message aggregation over neighborhood $\mathcal{N}(j)$:
\begin{equation}
    c_j^{t}=AGG(\{\phi(x_j^t, x_k^t)|k \in \mathcal{N}(j)\}),
\end{equation}
where $\phi(\cdot)$ is a learnable message function that produces neighbor messages. The deterministic transition is then calculated as follows:
\begin{equation}
    h_j^{t}=f_{dyn}(h_j^{t-1},z_j^{t-1},c_j^{t-1},A^{t-1}).
\end{equation}
We learn the node-wise recurrent state transition function $f_{dyn}$ with two distributions over the stochastic state: a posterior distribution $q(z_j^t|h_j^t,x_j^t,c_j^{t})$ relying on the current observation $(x_j^t,c_j^{t})$, and a prior distribution $p(\hat{z}_j^t|h_j^t)$ that depends only on the recurrent state. Two complementary KL regularizers are applied to align the prior and posterior. Specifically, the dynamic loss trains the prior distribution to match the inference posterior: 
\begin{equation}
    \mathcal{L}_{dyn}=\sum_t \sum_{j=1}^NKL[sg(q_\omega(z_j^t|h_j^t,x_j^t,c_j^t))||p_\omega(\hat{z}_j^t|h_j^t)],
\end{equation}

where $sg(\cdot)$ represents the stop gradient operator and $\omega$ is the dynamic model parameters. Conversely, the representation loss regularizes the posterior to remain consistent with the predictive dynamics along real trajectories:
\begin{equation}
    \mathcal{L}_{rep}=\sum_t \sum_{j=1}^NKL[q_\omega(z_j^t|h_j^t,x_j^t,c_j^t)||sg(p_\omega(\hat{z}_j^t|h_j^t))].
\end{equation}

Although $\mathcal{L}_{dyn}$ and $\mathcal{L}_{rep}$ regularize node-wise latent transitions by aligning the prior and posterior distributions, they do not constrain interaction-induced structural changes. Therefore, we introduce an auxiliary topology objective that targets such structural events, complementing the previous losses with direct structural supervision. Specifically, we predict the structural outcome $b^t$ of each interaction, and optimize

\begin{equation}
\mathcal{L}_{topo}(\rho)
= - \sum_t \log p_{\rho}\!\left(b^t \mid h_k^{t-1}, z_k^{t-1}, A^{t-1}\right),
\end{equation}

where $k$ denotes the node being interacted with at time $t-1$. We instantiate the structural outcome as node birth in the embodied domains (Appendix~\ref{app:graph_emb}) and the page reached after the interaction on the Web (Appendix~\ref{app:graph_web}).

\textbf{Dual Prediction.} 
To ensure that the imagined rollouts remain interpretable and predictive, the latent state should be decodable into a compact graph state. However, the previous objectives primarily regularize latent evolution and do not guarantee that the latent variables preserve task-relevant semantics. We therefore introduce a \emph{dual prediction loss} that supervises both (i) environment attributes and (ii) agent attributes. Concretely, given the node-wise latent state $h_j^t,z_j^t$, we predict the attributes of environment and agent nodes (Appendices~\ref{app:graph_emb} and~\ref{app:graph_web}). Denote environment node attributes as $o_j^t$ and agent attributes as $o_a^t$, the decoders produce $\hat{o}_j^t \sim p_{\psi_e}(o_j^t | h_j^t,z_j^t)$ and $\hat{o}_a^t \sim p_{\psi_a}(o_a^t| h_a^t,z_a^t)$, where $(h_a^t,z_a^t)$ are the agents' deterministic and stochastic latent states, $\psi_e$ and $\psi_a$ denote the learnable parameters of the environment node decoder and the agent decoder, respectively. We train the decoders by maximizing the likelihood of the corresponding targets, yielding:
\begin{equation}
    \mathcal{L}_{pred,e}(\psi_e) = \sum_t \sum_{j=1}^N-\log p_{\psi_e}(o_j^t|h_j^t,z_j^t),
\end{equation}

\begin{equation}
    \mathcal{L}_{pred,a}(\psi_a) = \sum_t -\log p_{\psi_a}(o_a^t|h_a^t,z_a^t).
\end{equation}
\begin{equation}
    \mathcal{L}_{pred} = \beta_e \mathcal{L}_{pred,e} + \beta_a \mathcal{L}_{pred,a} 
\end{equation}
Finally, the total loss is the weighted combination of the aforementioned losses:
\begin{equation}
    \mathcal{L}_{eg} = \sum_{q \in \mathcal{Q}}\lambda_q \mathcal{L}_{q},\label{eq:total}
\end{equation}
where $\mathcal{Q}=\{dyn, rep,topo,pred\}$, and $\lambda_q$ is the corresponding coefficient. In this way, we jointly stabilize node-wise latent dynamics, supervise interaction-induced structural events, and align node representations with their corresponding attributes in an end-to-end manner.

\subsection{Decision Making with RWM}
\label{sec:decision}
In embodied control, imagined graph rollouts train a policy over graph-level macro-actions. While in Web agents, the imagination helps a verifier compare candidate browser actions. The graph construction in each setting is given in Appendices~\ref{app:graph_emb} and~\ref{app:graph_web}.

\subsubsection{Embodied Control}
\label{sec:method:llwm}
In embodied control, the embodied graph is built from the segmented objects of the current observation, and an interaction is one of two types, moving to or acting on a target object (Appendix~\ref{app:graph_emb}).
Given the learned EG-RSSM and the imagination horizon $K$, we roll out graph latents $\{h^t, z^t\}_{t=t_s}^{t_s+K}$ from the posterior at time $t_s$. As in Dreamer~\cite{hafner2025dreamerv3}, a graph policy $\pi_{\text{g}}(A^t \mid h^t, z^t)$ is trained with actor-critic learning on these imagined trajectories to maximize the predicted return (details in Appendix~\ref{app:details:training}). Every $H$ environment steps, a macro-action $A^t=(i^t,v_{tar}^t)$ is selected under $\pi_{\text{g}}$. After that, an executor produces primitive actions $\{a_t,\ldots,a_{t+H-1}\}$ in the original environment action space to achieve the proposed macro-action $A^t$. It keeps the primitive action space unchanged and can be trained or replaced independently. In this case, we use a 200M-parameter DreamerV3 as the executor.

\subsubsection{Web Agents}
\label{sec:web_instantiation}
On webpages, the embodied graph is built from the DOM elements of the current page, and an interaction is a native browser action (click, fill, press, scroll, or select) on a target element (Appendix~\ref{app:graph_web}). At decision time, a frozen web agent, WebRL-Llama-3.1-8B~\cite{qi2025webrl}, proposes several candidate actions. From the common posterior, RWM starts the corresponding independent imagined branch with each candidate action. A verifier, Qwen2.5-VL-32B-Instruct~\cite{qwen2025qwen25vl}, receives the imagined rollouts to make the final decision.

\section{Experiments}
\label{sec:expri}

\subsection{Experiments on Embodied Control}
\label{sec:embodied_exp}

\subsubsection{Setup}
\label{sec:setup}
We evaluate on two long-horizon embodied benchmarks: \textbf{Minecraft} \citep{hill2023mineplanner} and \textbf{Crafter} \citep{hafner2021crafter}.

\textbf{Minecraft} is a procedurally generated open-world \emph{3D} sandbox where the agent must navigate large biomes and cave systems, manage survival signals (e.g., health and hunger), and interact with the environment by breaking blocks and fighting hostile mobs. We use \emph{diamond acquisition} as the target task, where agents need to explore, mine, and obtain diamonds under sparse and delayed rewards, and report the intermediate targets stone, iron and gold. The agent starts with the required tools so that it can focus on discovering and mining the targets. Following prior work \citep{hafner2025dreamerv3}, we increase the block breaking speed to accelerate training and form the observation and action space. Additionally, RWM obtains its graph candidates from the observed images with a learned object-segmentation front-end (Appendix~\ref{app:details:perception}).

\textbf{Crafter} \citep{hafner2021crafter} distills similar mechanics into a compact \emph{2D} setting in which the agent must gather materials, craft increasingly advanced tools, and handle hazards to unlock 22 achievements. Its setup is given in Appendix~\ref{app:crafter_achievements}.

\textbf{Baselines.} We compare with DreamerV3 \citep{hafner2025dreamerv3}, a state-of-the-art model-based agent that imagines trajectories in a latent space learned from high-dimensional observations and was the first to obtain diamonds in Minecraft without human data. LS-Imagine \citep{li2024lsimagine}, which augments imagination with long-horizon ``jumpy'' transitions for open-world exploration. And the model-free PPO \citep{schulman2017ppo} and Rainbow \citep{hessel2018rainbow}.

\textbf{Implementation details.} A macro-action is selected every $H=50$ environment steps. RWM has 4M parameters, and the executor is a 200M-parameter DreamerV3 \cite{hafner2025dreamerv3}. The complete hyperparameters are provided in Table~\ref{tab:app_hparams}. The computational complexity and measured computational cost are reported in Appendix~\ref{app:details:complexity}.

For Minecraft, all agents are trained for 5M environment steps with a maximum episode length of 36,000 steps, and we report results after 1M, 2M, 3M, 4M and 5M steps. For Crafter, we follow the default setting in \cite{hafner2021crafter}, training agents for 1M steps with a maximum episode length of 10,000. Unless otherwise stated, we apply the same environment configuration to all baselines. Results are the mean $\pm$ standard deviation over 3 training seeds.

\textbf{Metrics.} For Minecraft, we report the Success Rate (SR), the fraction of episodes in which the agent obtains the target item, and the Success Length (SL), the earliest time step at which the target is obtained, averaged over successful episodes (``--'' when no episode succeeds). For Crafter, we follow the standard evaluation protocol and report reward and the aggregated Crafter score, defined as the geometric mean over achievement success rates to emphasize broad and compositional skill acquisition.

\subsubsection{Results}
\label{sec:results}

\begin{table*}[t]
\centering
\caption{Minecraft results after 1M--5M environment steps: success rate (SR$\uparrow$, \%) and success length (SL$\downarrow$, steps, ``-- if none) are reported mean$\pm$std over 3 seeds. Best in bold, second best underlined. Ranks are computed among methods with non-zero SR.}
\label{tab:comparison}
\resizebox{.9\textwidth}{!}{
\begin{tabular}{cl|cc|cc|cc|cc}
\toprule
\multirow{2}{*}{\textbf{Steps}} & \multirow{2}{*}{\textbf{Model}} & \multicolumn{2}{c|}{\textbf{Stone}} & \multicolumn{2}{c|}{\textbf{Iron}} & \multicolumn{2}{c|}{\textbf{Gold}} & \multicolumn{2}{c}{\textbf{Diamond}} \\
\cmidrule(lr){3-4}\cmidrule(lr){5-6}\cmidrule(lr){7-8}\cmidrule(lr){9-10}
& & SR & SL & SR & SL & SR & SL & SR & SL \\
\midrule
\multirow{5}{*}{1M} & PPO & 92.40{\tiny$\pm$3.99} & 358{\tiny$\pm$8} & 28.76{\tiny$\pm$4.68} & 2628{\tiny$\pm$540} & 0.00{\tiny$\pm$0.00} & -- & 0.00{\tiny$\pm$0.00} & -- \\
 & Rainbow & 95.05{\tiny$\pm$0.70} & 663{\tiny$\pm$324} & 29.80{\tiny$\pm$3.77} & 2320{\tiny$\pm$635} & 0.00{\tiny$\pm$0.00} & -- & 0.00{\tiny$\pm$0.00} & -- \\
 & DreamerV3 & \textbf{97.03{\tiny$\pm$0.80}} & \textbf{264{\tiny$\pm$18}} & \underline{30.82{\tiny$\pm$4.66}} & \textbf{1875{\tiny$\pm$255}} & \underline{2.58{\tiny$\pm$2.21}} & \textbf{1190{\tiny$\pm$92}} & 0.00{\tiny$\pm$0.00} & -- \\
 & LS-Imagine & \underline{96.41{\tiny$\pm$1.78}} & \underline{279{\tiny$\pm$78}} & 25.11{\tiny$\pm$4.64} & \underline{2103{\tiny$\pm$184}} & 1.89{\tiny$\pm$1.37} & 5522{\tiny$\pm$721} & 0.00{\tiny$\pm$0.00} & -- \\
 & \textbf{RWM (Ours)} & 95.57{\tiny$\pm$2.76} & 420{\tiny$\pm$138} & \textbf{32.89{\tiny$\pm$4.62}} & 2115{\tiny$\pm$113} & \textbf{4.19{\tiny$\pm$0.53}} & \underline{3054{\tiny$\pm$1351}} & \textbf{1.47{\tiny$\pm$1.11}} & \textbf{8198{\tiny$\pm$708}} \\
\midrule
\multirow{5}{*}{2M} & PPO & 93.76{\tiny$\pm$0.13} & 405{\tiny$\pm$4} & 29.18{\tiny$\pm$3.38} & 2317{\tiny$\pm$405} & 0.00{\tiny$\pm$0.00} & -- & 0.00{\tiny$\pm$0.00} & -- \\
 & Rainbow & 95.40{\tiny$\pm$0.71} & 556{\tiny$\pm$246} & 32.55{\tiny$\pm$3.85} & 2294{\tiny$\pm$134} & 0.00{\tiny$\pm$0.00} & -- & 0.00{\tiny$\pm$0.00} & -- \\
 & DreamerV3 & \underline{97.12{\tiny$\pm$0.99}} & \textbf{197{\tiny$\pm$12}} & 34.01{\tiny$\pm$1.54} & 2385{\tiny$\pm$653} & 2.62{\tiny$\pm$0.68} & 5393{\tiny$\pm$4403} & 0.00{\tiny$\pm$0.00} & -- \\
 & LS-Imagine & \textbf{97.32{\tiny$\pm$0.84}} & 402{\tiny$\pm$23} & \underline{35.72{\tiny$\pm$1.75}} & \textbf{1273{\tiny$\pm$381}} & \underline{2.85{\tiny$\pm$1.08}} & \underline{3190{\tiny$\pm$2311}} & 0.00{\tiny$\pm$0.00} & -- \\
 & \textbf{RWM (Ours)} & 96.51{\tiny$\pm$0.69} & \underline{337{\tiny$\pm$34}} & \textbf{38.28{\tiny$\pm$1.96}} & \underline{1647{\tiny$\pm$110}} & \textbf{3.13{\tiny$\pm$1.49}} & \textbf{2798{\tiny$\pm$1196}} & \textbf{1.55{\tiny$\pm$1.12}} & \textbf{5401{\tiny$\pm$297}} \\
\midrule
\multirow{5}{*}{3M} & PPO & 94.14{\tiny$\pm$0.08} & 363{\tiny$\pm$1} & 29.05{\tiny$\pm$3.28} & 2533{\tiny$\pm$183} & 0.00{\tiny$\pm$0.00} & -- & 0.00{\tiny$\pm$0.00} & -- \\
 & Rainbow & 95.44{\tiny$\pm$0.65} & 529{\tiny$\pm$139} & 33.19{\tiny$\pm$3.12} & 2126{\tiny$\pm$274} & 0.00{\tiny$\pm$0.00} & -- & 0.00{\tiny$\pm$0.00} & -- \\
 & DreamerV3 & \textbf{98.21{\tiny$\pm$0.59}} & \textbf{170{\tiny$\pm$9}} & 38.63{\tiny$\pm$1.11} & 2265{\tiny$\pm$501} & 1.61{\tiny$\pm$0.78} & 5534{\tiny$\pm$4311} & 0.00{\tiny$\pm$0.00} & -- \\
 & LS-Imagine & 97.17{\tiny$\pm$0.66} & \underline{233{\tiny$\pm$47}} & \underline{38.77{\tiny$\pm$1.68}} & \textbf{1428{\tiny$\pm$444}} & \underline{3.24{\tiny$\pm$1.10}} & \textbf{3775{\tiny$\pm$1774}} & 0.00{\tiny$\pm$0.00} & -- \\
 & \textbf{RWM (Ours)} & \underline{97.77{\tiny$\pm$0.73}} & 292{\tiny$\pm$85} & \textbf{48.61{\tiny$\pm$1.25}} & \underline{1746{\tiny$\pm$387}} & \textbf{3.63{\tiny$\pm$1.06}} & \underline{4021{\tiny$\pm$2237}} & \textbf{2.97{\tiny$\pm$1.80}} & \textbf{5306{\tiny$\pm$236}} \\
\midrule
\multirow{5}{*}{4M} & PPO & 88.17{\tiny$\pm$5.67} & 548{\tiny$\pm$229} & 29.10{\tiny$\pm$5.38} & 2281{\tiny$\pm$572} & 0.00{\tiny$\pm$0.00} & -- & 0.00{\tiny$\pm$0.00} & -- \\
 & Rainbow & 95.48{\tiny$\pm$1.17} & 568{\tiny$\pm$227} & 33.89{\tiny$\pm$4.18} & 2269{\tiny$\pm$144} & 0.00{\tiny$\pm$0.00} & -- & 0.00{\tiny$\pm$0.00} & -- \\
 & DreamerV3 & \underline{98.44{\tiny$\pm$0.26}} & \textbf{157{\tiny$\pm$12}} & 39.51{\tiny$\pm$1.62} & \underline{2075{\tiny$\pm$108}} & 1.39{\tiny$\pm$0.93} & 5941{\tiny$\pm$4098} & 0.00{\tiny$\pm$0.00} & -- \\
 & LS-Imagine & \textbf{98.62{\tiny$\pm$0.30}} & \underline{251{\tiny$\pm$29}} & \underline{45.14{\tiny$\pm$1.59}} & 3006{\tiny$\pm$386} & \underline{2.81{\tiny$\pm$1.20}} & \underline{4218{\tiny$\pm$710}} & 0.00{\tiny$\pm$0.00} & -- \\
 & \textbf{RWM (Ours)} & 97.49{\tiny$\pm$0.97} & 351{\tiny$\pm$45} & \textbf{47.96{\tiny$\pm$1.56}} & \textbf{2022{\tiny$\pm$664}} & \textbf{3.92{\tiny$\pm$0.80}} & \textbf{3809{\tiny$\pm$2321}} & \textbf{5.72{\tiny$\pm$1.78}} & \textbf{4740{\tiny$\pm$295}} \\
\midrule
\multirow{5}{*}{5M} & PPO & 85.64{\tiny$\pm$5.26} & 790{\tiny$\pm$258} & 25.36{\tiny$\pm$8.18} & 2303{\tiny$\pm$476} & 0.00{\tiny$\pm$0.00} & -- & 0.00{\tiny$\pm$0.00} & -- \\
 & Rainbow & 95.77{\tiny$\pm$1.06} & 364{\tiny$\pm$86} & 34.16{\tiny$\pm$2.96} & 2316{\tiny$\pm$394} & 0.00{\tiny$\pm$0.00} & -- & 0.00{\tiny$\pm$0.00} & -- \\
 & DreamerV3 & \underline{98.36{\tiny$\pm$0.34}} & \textbf{136{\tiny$\pm$11}} & 40.76{\tiny$\pm$0.83} & 2252{\tiny$\pm$108} & 1.99{\tiny$\pm$1.07} & 6352{\tiny$\pm$5316} & 0.00{\tiny$\pm$0.00} & -- \\
 & LS-Imagine & 98.23{\tiny$\pm$0.79} & \underline{283{\tiny$\pm$36}} & \underline{43.88{\tiny$\pm$1.79}} & \underline{2114{\tiny$\pm$101}} & \underline{2.75{\tiny$\pm$1.46}} & \underline{5684{\tiny$\pm$3860}} & 0.00{\tiny$\pm$0.00} & -- \\
 & \textbf{RWM (Ours)} & \textbf{98.75{\tiny$\pm$0.89}} & 313{\tiny$\pm$62} & \textbf{48.73{\tiny$\pm$1.75}} & \textbf{1582{\tiny$\pm$93}} & \textbf{3.52{\tiny$\pm$0.86}} & \textbf{3962{\tiny$\pm$2404}} & \textbf{6.94{\tiny$\pm$2.12}} & \textbf{4512{\tiny$\pm$1014}} \\
\bottomrule
\end{tabular}
}
\end{table*}

Table~\ref{tab:comparison} shows that RWM outperforms the baselines, with the largest gains on long-horizon targets.

For relatively simple tasks, such as collecting stone and iron, their performances remain similar since these tasks can be solved through simple random exploration without requiring a deep understanding of environment dynamics. The performance gap becomes more evident on challenging tasks, such as collecting gold and diamonds. These materials are buried deeper and associated with sparse rewards. Among the strong baselines, RWM outperforms the others by capturing abstract object relations and leveraging sparse feedback more effectively, resulting in continued gains and shorter average success lengths. Notably, it is the only method that finds diamonds within the given 5M training budget.

To further assess the generality of our approach beyond \emph{3D} scenarios, we evaluate RWM on Crafter. The results are provided in Table \ref{tab:crafter}, which exhibit a similar trend to those on Minecraft. Compared with the baselines, RWM achieves the best performance in terms of both Crafter score and episodic reward. Since the Crafter score is defined as the geometric mean over achievement success rates \cite{hafner2021crafter}, the improvement in both reward and score suggests that RWM not only improves the overall episode performance, but also helps the agent make broader progress across diverse achievements. A further analysis of the learned agent skills is provided in Appendix~\ref{app:crafter_achievements}, where RWM achieves comparable performance on simple early-stage achievements and obtains clear gains on harder long-horizon achievements that require multi-step resource collection and crafting.

\begin{table}[htbp!]
\centering
\caption{Ablation study in the Minecraft diamond task. Variant 1: w/o admission scorer. Variant 2: w/o relational propagation. Success rate (SR, \%) and success length (SL, steps) are reported by mean$\pm$std over 3 seeds.}
\label{tab:ablation_results}
\setlength{\tabcolsep}{3.5pt}
\renewcommand{\arraystretch}{1.05}
\resizebox{.95\linewidth}{!}{
\begin{tabular}{llccccc}
\toprule
Model & Metric & 1M & 2M & 3M & 4M & 5M \\
\midrule

\multirow{2}{*}{RWM}
& SR   & \textbf{1.47{\tiny$\pm$1.11}}     & \textbf{1.55{\tiny$\pm$1.12}}   & \textbf{2.97{\tiny$\pm$1.80}}   & \textbf{5.72{\tiny$\pm$1.78}}   & \textbf{6.94{\tiny$\pm$2.12}}  \\
& SL  & \textbf{8198{\tiny$\pm$708}}    & \textbf{5401{\tiny$\pm$297}}   & \textbf{5306{\tiny$\pm$236}}  & \textbf{4740{\tiny$\pm$295}}   & 4512{\tiny$\pm$1014}  \\

\midrule
\multirow{2}{*}{Variant 1}
& SR   & 0.00{\tiny$\pm$0.00}     & 0.00{\tiny$\pm$0.00}     & 0.00{\tiny$\pm$0.00}     & 0.00{\tiny$\pm$0.00}     & 0.00{\tiny$\pm$0.00} \\
& SL  & --    & --    & --    & --    & -- \\

\midrule
\multirow{2}{*}{Variant 2}
& SR   & 0.00{\tiny$\pm$0.00}     & 1.33{\tiny$\pm$0.62} & 1.79{\tiny$\pm$0.71} & 1.66{\tiny$\pm$0.69}  & 2.48{\tiny$\pm$0.95} \\
& SL  & --    & 6750{\tiny$\pm$294}  & 5730{\tiny$\pm$201}  & 5971{\tiny$\pm$337}  & \textbf{3945{\tiny$\pm$883}} \\

\bottomrule
\end{tabular}}
\end{table}

\begin{table}[t]
\centering
\caption{Experimental results in Crafter after 1M environment steps (mean$\pm$std).}
\label{tab:crafter}
\resizebox{.95\linewidth}{!}{
\begin{tabular}{lcccc}
\toprule
\textbf{Model} & \textbf{PPO} & \textbf{Rainbow} & \textbf{DreamerV3} & \textbf{RWM}\\
\midrule
Score (\%) & 4.6$\pm$0.3 & 4.3$\pm$0.2 & 14.5$\pm$1.6 & \textbf{15.8$\pm$2.0} \\
Reward  & 4.2$\pm$1.2 & 6.0$\pm$1.3 & 11.7$\pm$1.9 &  \textbf{12.2$\pm$1.7} \\
\bottomrule
\end{tabular}}
\end{table}

\subsubsection{Imagination Drift Analysis}
\label{sec:drift}
\begin{figure}
\vspace{20pt}
    \centering
    \includegraphics[width=\linewidth]{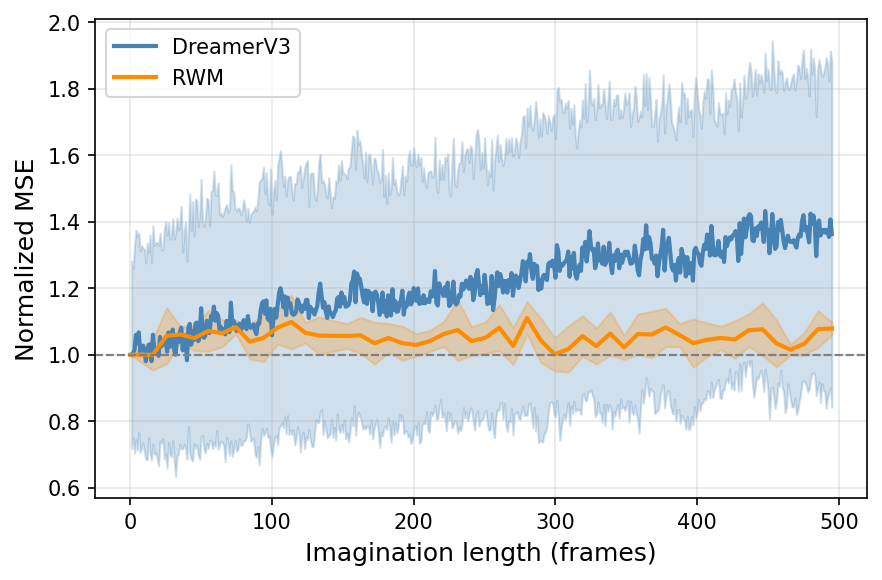}
    \caption{Open-loop imagination drift under different imagination lengths. Compared with DreamerV3, RWM maintains a nearly stable prediction error over long rollouts, showing that our proposed method effectively reduces compounding imagination drift.}
    \label{fig:imaginedrift}
\end{figure}

To directly examine whether RWM mitigates compounding prediction errors during long-horizon imagination, we evaluate the accumulating imagination drift in Fig. \ref{fig:imaginedrift}. Starting from the same posterior state, each model performs open-loop imagination for up to 500 environment steps, and we measure the discrepancy between the imagined state and the corresponding ground-truth future state through the decoded predictions. Specifically, since the decoders and prediction targets vary across methods, the raw MSE values are not directly comparable. For a fair comparison, we normalize the MSE at each imagination step by the MSE at the first imagination step (Normalized MSE), so that the metric reflects the relative growth of prediction error, i.e., the magnitude of imagination drift. As shown in Fig. \ref{fig:imaginedrift}, the drift increases as the imagination horizon becomes longer, confirming that long-horizon rollout is inherently vulnerable to error accumulation. However, RWM exhibits much slower drift growth than the dense-observation baseline (DreamerV3). This indicates that the proposed embodied graph abstraction and relation-aware EG-RSSM provide a more stable predictive structure, thereby extending the effective imagination horizon and supporting more reliable macro-action selection. In addition, the smaller variance of RWM further suggests that our method is more stable across rollout samples.

This is consistent with where RWM helps in Table~\ref{tab:comparison}: short rollouts suffice for simple targets, whereas diamond acquisition requires coordinating many interdependent interactions, where small rollout errors lead to wrong macro-action choices.

\subsubsection{Ablation Study}
\label{sec:abl}
To isolate the contributions of key design choices in \textbf{RWM}, we conduct ablation studies by constructing two variants:
\begin{enumerate}[label=\textbf{(\arabic*)}]
    \item \textbf{w/o admission scorer in TA-EGC.} We remove the admission scorer in task-aware embodied graph constructor. Instead of selecting task-relevant nodes, we update the embodied graph with all nodes extracted from the local observation at every step. This variant tests whether the proposed sparse, task-aware embodied graph constructor is necessary to filter redundancy and preserve the informative structure. 
    \item \textbf{w/o relational propagation in EG-RSSM.} We remove relational propagation in EG-RSSM by disabling message passing across edges such that each node latent is updated independently without relational context. This variant keeps the same object-level graph state, constructor, and node-wise latent dynamics, and removes only the relations. It therefore separates the value of explicit relations from the value of moving from pixels to an object-level state. 
\end{enumerate}

We evaluate the two ablations on the diamond task. As shown in Table~\ref{tab:ablation_results}, removing either component consistently degrades performance, indicating that both of them are crucial for RWM. Variant 1 (without the task-aware admission scorer) completely fails to obtain the diamond. This suggests that without controlled admissions, embodied graph may becomes unstable and changes frequently, making it difficult to learn consistent transition dynamics. Variant 2 (without relational propagation) exhibits a substantially lower SR, indicating that relational context contributes beyond the object-level state representation. The gap between Variant 2 and the full model widens as training progresses. Their mean success rates are close at 2M steps, but by 4M steps, the full model reaches 5.72\%, while Variant 2 remains at 1.66\%. At 5M steps, their success rates reach 6.94\% and 2.48\%, respectively. Both models can discover successful trajectories, but the full model shows larger improvements with continued training. This pattern suggests that relational propagation contributes beyond initial task discovery, helping the agent make better use of additional training experience.

\subsection{Evaluation on Interactive Webpages}
\label{sec:web_evaluation}

Beyond embodied control, we test whether the same relational world model carries over to Web agents, where page elements and their DOM relations form the graph.

\subsubsection{Benchmark.} 
We evaluate 35 content-management tasks (CMS35) of WebArena-Lite~\cite{liu2025visualagentbench,zhou2024webarena}, which are performed on the administration back end of an e-commerce website (Magento). Each episode starts from a freshly deployed site and is limited to 30 steps. We use the official observation and action spaces and the official evaluators.

\subsubsection{Training data.} 
RWM is trained offline on 94 recorded browser trajectories collected by a frozen A3-Qwen3.5-9B Web agent~\cite{lu2026a3} on 12 WebArena CMS tasks. None of these tasks, nor their task templates appears in CMS35.

\subsubsection{Baselines.} 
We compare RWM with a pixel-based world model under the decision procedure of Section~\ref{sec:web_instantiation}: WebRL-Llama-3.1-8B~\cite{qi2025webrl} proposes the candidate actions, a world model forecasts the consequence of each candidate, and another verifier (Qwen2.5-VL-32B-Instruct~\cite{qwen2025qwen25vl}) selects one of them. The three variants differ only in what the verifier receives for each candidate. Verifier-only branch receives the candidates without any forecast. Verifier + image RSSM additionally receives the next screen predicted by a DreamerV3-style pixel world model on 288×512 screenshots, trained on the same trajectories with the same number of updates and history length. Verifier + RWM receives the element-level effects predicted by RWM. Each task is run once per variant from a freshly deployed site. For reference, we also report the published WebArena-Lite results of WebRL and other agents.

\subsubsection{Decision-making on CMS35.} \label{sec:web_results} 
Table~\ref{tab:web_main} reports task success on CMS35. Without forecasts, the verifier-only branch solves 45.7\% of the tasks, even worse than the pure WebRL-Llama-3.1-8B. Pixel forecasts raise this to 51.4\%, and RWM further improves it to 60.0\%. This indicates that without proper evidence on the consequences of the candidate actions, a general LLM struggles to tell which action is correct. The predicted screen conveys the consequence of an action only implicitly. Changes in the state of a single element are hard to read from it. Whereas RWM explicitly indicates which elements change and how, which the verifier can relate directly to the task. 

\subsubsection{Case Study}
\label{app:web_case}
We further examine some cases to verify how RWM takes effect to increase the success rate and how other methods fails.

\begin{figure}[t]
\centering
\includegraphics[width=\columnwidth]{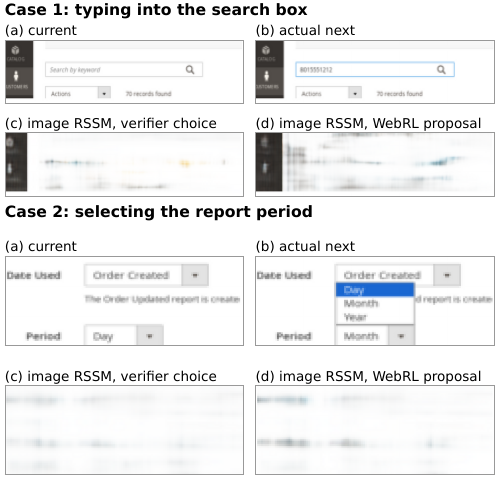}
\caption{Case study on CMS35. Top: Case 1, typing a phone number into the search box. Bottom: Case 2, selecting the report period. Each case is enlarged around the changed element and shows (a) the current screen, (b) the actual next screen after the verifier's choice, (c) the image RSSM's prediction for the verifier's choice, and (d) its prediction for WebRL's proposal. RWM forecasts the new values exactly (\texttt{8015551212} in the search box and \texttt{month} in the Period drop-down), whereas the image RSSM's predictions are blurred and nearly identical for the two candidates.}
\label{fig:web_case}
\end{figure}

\textbf{Case 1: searching customers.} The task is ``Find the customer name and email with phone number 8015551212'', and the agent is on the customer list (Fig.~\ref{fig:web_case}). WebRL proposes the first action to click the \emph{Filters} button, for which RWM forecasts that the search box stays empty. Therefore, the verifier instead selects typing \texttt{8015551212} into the \emph{Search by keyword} box, for which RWM forecasts that the box receives focus and its value becomes \texttt{8015551212}. This helps the RWM branch to success in this task. Instead, the image RSSM predicts a blurred page for both candidates. The search box shows no text, and the two predicted screens are nearly indistinguishable. This makes the verifier fails to choose the right action.

\textbf{Case 2: configuring a report.} The task asks for the monthly count of successful orders in 2022, and the agent is on the Orders Report page. WebRL proposes a wrong action to type \texttt{12/31/2022} into the \emph{From} field. The verifier instead selects \texttt{month} in the \emph{Period} drop-down, for which RWM forecasts that Period changes from \texttt{day} to \texttt{month} with all other fields unchanged, again matching the actual next state. The predicted screens of the image RSSM do not show this change and are again nearly identical across the two candidates.

In both cases the initial WebRL proposal is not good enough, and the action consequence is a field value that covers a small portion of the screen, which a pixel forecast does not resolve but an element-level forecast states directly. And this is how RWM help with the higher success rate.

\begin{table}[t]
\centering
\caption{Success rate (\%) on WebArena-Lite CMS35. Upper block: published results~\cite{qi2025webrl}. Lower block: our runs, in which WebRL-Llama-3.1-8B proposes the candidates and the same verifier selects one. The variants differ only in the forecasts given to the verifier.}
\label{tab:web_main}
\begin{tabular}{lc}
\toprule
Method & CMS SR \\
\midrule
GPT-4-Turbo & 14.3 \\
GPT-4o & 20.0 \\
AutoWebGLM & 28.6 \\
Llama3.1-8B + DigiRL & 37.1 \\
GLM-4-9B + WebRL & 48.6 \\
WebRL-Llama-3.1-8B & 54.3 \\
WebRL-Llama-3.1-70B & 54.3 \\
\midrule
Verifier (no forecast) & 45.7 \\
Verifier + image RSSM & 51.4 \\
Verifier + RWM & \textbf{60.0} \\
\bottomrule
\end{tabular}
\end{table}

\section{Related Work}
\label{sec:rw}

\textbf{World Models.}
World models learn internal representations~\citep{bruce2024genie,chen2022transdreamer,robine2023transformer,wang2024worlddreamer} and predictive dynamics~\citep{hafner2020dreamer,hafner2021dreamerv2,okada2022dreamingv2,wu2023daydreamer, Zhan_2026_CVPR} to enable imagination-based reasoning and planning. Early work~\citep{ha2018worldmodels} simulates future observations with latent generative RNNs, while recent methods scale capacity and generalization. PWM~\citep{georgiev2024pwm} leverages large pre-trained world models for policy learning across continuous-control tasks, STORM~\citep{zhang2023storm} combines stochastic latent dynamics with transformers for strong multi-step prediction on Atari, and Dreamer-style agents~\citep{hafner2020dreamer,hafner2021dreamerv2,hafner2025dreamerv3} learn from imagined trajectories for sample-efficient long-horizon credit assignment. LS-Imagine~\citep{li2024lsimagine} further extends imagination in open-world settings via long-horizon ``jumpy'' transitions. Object-centric world models, such as C-SWM~\citep{kipf2020contrastivelearningstructuredworld}, OP3~\citep{veerapaneni2020op3} and SlotFormer~\citep{wu2023slotformer}, factorize the state into object slots discovered from pixels, mostly in small synthetic scenes with a fixed number of objects. Most world models still rely on dense pixel-level or entangled latent representations and lack explicit relational structure, which can limit rollout stability and the effective imagination horizon. Open-world tasks further require the set of modeled objects to change online as the agent explores.

\textbf{Graphs in Embodied AI.}
Graphs provide structured representations of entities and relations for embodied perception and reasoning. Scene-graph methods such as SG-Nav~\cite{yin2024sgnav}, ESCA~\cite{huang2025esca}, and GraphEQA~\cite{saxena2025GraphEQA} support navigation, grounding, and question answering with 3D semantic context. EmbodiedVSR~\citep{zhang2025embodiedvsr} updates a dynamic scene graph after each action to guide the chain-of-thought reasoning of a multimodal language model, but the graph serves as input for reasoning rather than as the state of a learned dynamics model. Graphs are also used as supervision, e.g., Scene Graph Contrastive loss~\citep{singh2022generalpurposesupervisorysignal}, and as knowledge memories for instruction following, e.g., Optimus-1~\cite{li2025optimus} and ExRAP~\cite{yoo2025exploratoryretrievalaugmentedplanningcontinual}. In model-based and hierarchical RL, graph abstractions have been used for planning with skills, including Deep Skill Graphs~\cite{pmlr-v139-bagaria21a} and intrinsically motivated temporally abstract graph-based world models~\citep{bagaria2025intrinsically}. In these methods, nodes represent subgoals or abstract states, and edges represent skills that move the agent between them through sequences of primitive actions. The graph records which subgoals can be reached through skill execution, allowing the planner to combine several skills into a longer plan. Despite these advances, these methods primarily use graphs for perception, reasoning, or skill-level planning, rather than modeling interaction-driven changes in entity states over long imagined rollouts.

\textbf{World Models for Web Agents.}
Web agents act on realistic websites through browser actions such as clicking, typing and selecting. WebArena~\cite{zhou2024webarena} provides self-hosted websites for e-commerce, content management, forums, code hosting and maps, and WebArena-Lite~\cite{liu2025visualagentbench} is a human-verified subset of its tasks. WebRL~\cite{qi2025webrl} trains LLM agents on WebArena-Lite with online reinforcement learning, using a self-evolving curriculum and an outcome-supervised reward model. To anticipate the consequences of actions before committing to them, recent work equips Web agents with world models. WMA~\cite{chae2025webagentsworldmodels} trains an LLM to predict the change of the observation caused by an action in natural language and uses these predictions to select actions at inference time, and WebDreamer~\cite{gu2025webdreamer} prompts an LLM to simulate the outcomes of candidate actions and plans by scoring the simulated outcomes. However, these world models describe the next state as free-form text and do not explicitly represent page elements and their relations, which makes it difficult to tell which elements an action changes and to verify a prediction against the actual page.

\section{Limitations}
\label{sec:limitations}
RWM relies on an entity-level front-end: object-segmentation models trained with object labels in Minecraft and Crafter, and the DOM on the Web. The result may be affected by the quality of the front-end, and transfer to real-world settings requires a detector for the relevant objects. Moreover, predefined relations ($K_{nn}$ spatial neighbors, DOM links) are adopted as edges, which may be unavailable in some cases. Learning the edges is left for future works. 

\section{Conclusion}
\label{sec:conlcusion}
Existing world models often struggle with long-horizon tasks because prediction errors accumulate during imagination, leading to drift in long rollouts. We proposed the \textbf{Relational World Model (RWM)}, a graph world model that imagines over \textbf{Embodied Graph}, a sparse relational state abstraction. RWM combines a \emph{Task-Aware Embodied Graph Constructor} with an \emph{Embodied Graph Recurrent State Space Model}, and is instantiated for embodied control and Web agents. In Minecraft, RWM's open-loop prediction error grows much more slowly than that of a pixel world model, and the resulting agent is the only one in our comparison that obtains diamonds within the given budget. On webpages, RWM forecasts element-level action consequences. Given these forecasts, a verifier selects better actions, and solves more tasks on WebArena-Lite. These findings suggest that sparse relational abstractions are a useful way to extend the effective horizon of world models, and their value grows with the difficulty of the task.

\clearpage
\bibliographystyle{ACM-Reference-Format}
\bibliography{main}

\appendix
\section{Implementation Details and More Results on Embodied Control}
\label{app:impledetails}

\subsection{Graph Construction}
\label{app:graph_emb}
In Minecraft and Crafter, candidates are produced by an object-segmentation front-end applied to the image observation (Appendix~\ref{app:details:perception}). An interaction is one of two types, moving to or acting on a target object. The graph has a fixed capacity of $N$ nodes, where node zero is the agent. Each object node carries its position $p\in\mathbb{R}^3$, type $\tau$, size $m$ and last-seen step $l$, and the agent node carries its position and inventory. A candidate matches an existing node if both have the same type and their squared distance is below $\delta^2$, and the closest such candidate is selected. Edges connect each node to its $K_{nn}$ nearest neighbors in space. The context $c_j^t$ of EG-RSSM is computed by an MPNN over these edges with the same number of layers and hidden size as the dynamics model. All scorer labels are computed from the front-end's detections and the agent's own decisions. A candidate receives a positive task label if it lies within a distance threshold of the target of a macro-action chosen within the next $W$ steps, and the task loss is weighted by the policy advantage. Novelty and coverage labels use type rarity and distance from the agent. The prediction heads of EG-RSSM decode node position, type and size, and the agent's position and inventory. The structural outcome in $\mathcal{L}_{topo}$ is node birth, a binary indicator of whether the interaction creates new nodes.

\subsection{Perception Front-End and Node Candidates}
\label{app:details:perception}
\textbf{Minecraft.} The front-end has two branches applied to the same frame. An instance branch, a YOLOv8m segmentation model~\cite{jocher2023yolov8}, detects and segments objects in nine categories: log, tree, crafting table, furnace, gold ore, diamond ore, iron ore, coal ore and zombie. A semantic branch, DeepLabV3+~\cite{chen2018deeplabv3plus} with a ResNet-50 encoder~\cite{he2016resnet}, labels every pixel as background, water, grass, sand, lava or stone. Both branches are trained on frames collected from Minecraft episodes with random exploration, which we annotated manually with the category labels above, and are frozen during RL. Instance masks take precedence over semantic labels where they overlap. Each instance yields a candidate with its type, mask centroid, bounding box and area. The 2D outputs are lifted to 3D with a monocular depth model (Video Depth Anything~\cite{chen2025videodepthanything}), which predicts depth. The 3D position is used to match candidates across steps.

\begin{figure}[t]
  \centering
  \includegraphics[width=\linewidth]{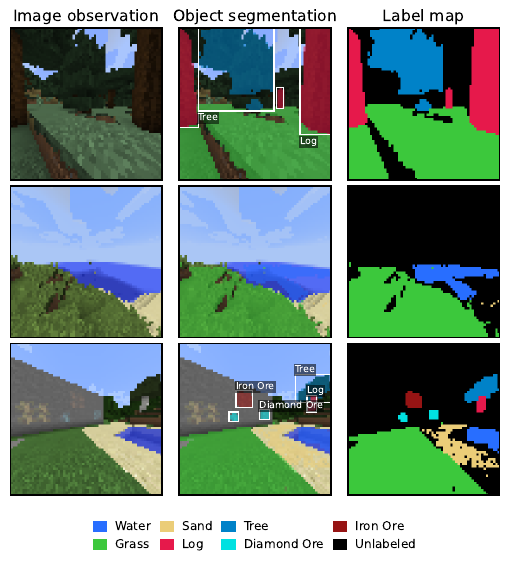}
  \caption{Perception front-end in Minecraft. Each row corresponds to a $64\times64$ observation (left), the segmentation (middle), and the resulting label map (right).}
  \label{fig:mc_frontend}
\end{figure}

Figure~\ref{fig:mc_frontend} shows the front-end output on the agent's $64\times64$ observations in three scenes. Large regions such as grass, water and sand are segmented as contiguous areas, and small objects such as single ore blocks are kept as separate instances. The areas and instantces form the node candidates.

\textbf{Crafter.} As the observation image in Crafter is a grid of different objects, segmentation is naturally provided. Node candidates are produced by an object semantic classifier MLP head trained on Crafter observations.

\subsection{Observation and Action space}
\label{app:details:encoding}
For Minecraft and Crafter, we follow the observation and action spaces of DreamerV3~\cite{hafner2025dreamerv3}. 

\subsection{Goal-Conditioned Executor and Macro-Action}
\label{app:details:lowlevel}
The executor is a 200M-parameter DreamerV3. It takes dense observations together with the macro-action and outputs primitive actions. The macro-action consists of an interaction type and a target node. The executor receives the interaction type and the target's position as its goal, which is embedded and concatenated with the policy inputs, without modifying the observation or action spaces. A maximum of $H=50$ environment steps are given to the executor to achieve the goal.

\subsection{Optimization}
\label{app:details:training}
RWM is trained with the total loss of Eq.~\eqref{eq:total} (weights in Table~\ref{tab:app_hparams}) using DreamerV3's optimizer settings unless otherwise stated. For policy learning in embodied control, we summarize the graph by a readout $g^t$ that concatenates the latent state of the agent node with pooled latent states of the object nodes. Reward and continuation heads predict $\hat{r}^t$ and $\hat{c}^t$ from $g^t$. The graph policy assigns a score to each admissible interaction based on the latent state of its target node and selects an interaction by a softmax over these scores. A value function on $g^t$ is trained on $\lambda$-return targets, as in DreamerV3.

\subsection{Online Update Cost and Graph Size}
\label{app:details:complexity}
The graph has fixed capacity, so the per-step cost and memory of RWM do not grow with episode length. Only the current graph $G^t$ is kept online (at most $N$ nodes and $K_{nn}N$ directed edges. Each step of TA-EGC has four parts. First, three small MLP heads score at most $M$ candidates, which costs $O(M)$. Second, the Top-$P$ candidates are selected in $O(M\log P)$. Third, each of the $P$ admitted candidates is compared with the $N$ existing nodes of the same type, which costs $O(PN)$, and the nodes unseen for the longest time are evicted when the graph is full. Finally, the $K_{nn}$-nearest-neighbor edges are rebuilt. Our implementation computes all pairwise distances for this, which costs $O(N^2)$ and is cheap for $N=128$. Additionally, EG-RSSM's message passing costs $O((N+K_{nn}N)\,d)$ per step for hidden size $d$. In Minecraft ($N=128$, $M=256$, $K_{nn}=4$) the full agent runs at 63.28$\pm$0.88 policy steps per second, compared with 70.76$\pm$1.53 for DreamerV3 with the same hyperparameters (AMD EPYC 7543 CPU, single NVIDIA A100 GPU).

\begin{table*}[t]
\centering
\caption{Key hyperparameters for RWM. Unless otherwise specified, shared hyperparameters follow the default DreamerV3 configuration~\cite{hafner2025dreamerv3}.}
\label{tab:app_hparams}
\small
\begin{tabular}{lc|lc}
\toprule
Hyperparameter & Value & Hyperparameter & Value \\
\midrule
Macro-action decision interval ($H$) & 50 & Decoder layers & 3 \\
Max nodes in the embodied graph ($N$) & 128 & Policy layers & 2 \\
Max observation candidates ($M$) & 256 & Value layers & 3 \\
node matching threshold ($\delta$) & 2.0 & Imagination length ($K$) & 15 \\
KNN neighbors ($K_{nn}$) & 4 & Loss scale (dynamics) ($\lambda_{\mathrm{dyn}}$) & 1.0 \\
Replay size & $5\times 10^5$ & Loss scale (representation) ($\lambda_{\mathrm{rep}}$) & 0.1 \\
Batch size & 16 & Loss scale (topology) ($\lambda_{\mathrm{topo}}$) & 1.0 \\
Batch length & 16 & Loss scale (prediction) ($\lambda_{\mathrm{pred}}$) & 1.0 \\
Train ratio & 1 & Loss scale (reward) ($\lambda_{\mathrm{rew}}$) & 1.0 \\
Learning rate & $4\times 10^{-5}$ & Loss scale (policy) ($\lambda_{\pi}$) & 1.0 \\
Deterministic state dim & 512 & Loss scale (value) ($\lambda_{V}$) & 1.0 \\
RSSM hidden dim & 64 & Score scale (task) ($\alpha_{t}$) & 1.0 \\
Stochastic state dim & 32 & Score scale (novel) ($\alpha_{n}$) & 1.0 \\
Discrete classes (per stoch dim) & 4 & Score scale (cover) ($\alpha_{c}$) & 0.5 \\
RSSM blocks & 4 & Loss scale (environment nodes) ($\beta_{e}$) & 1.0 \\
Hidden dimensions (encoder, decoder, reward) & 512 & Loss scale (agent node) ($\beta_{a}$) & 1.0 \\
Encoder layers & 2 &  &  \\
\bottomrule
\end{tabular}
\end{table*}

\subsection{Crafter Setup and More Results}
\label{app:crafter_achievements}
\textbf{Setup.} We use the standard Crafter environment~\cite{hafner2021crafter} with $64\times64$ image observations, the 17 discrete actions, and the default episode length of 10,000 steps. We train every agent for 1M environment steps. The macro-action interval, graph capacity and other RWM hyperparameters are provided in Table~\ref{tab:app_hparams}.

\begin{figure*}[t]
    \centering
    \includegraphics[width=\linewidth]{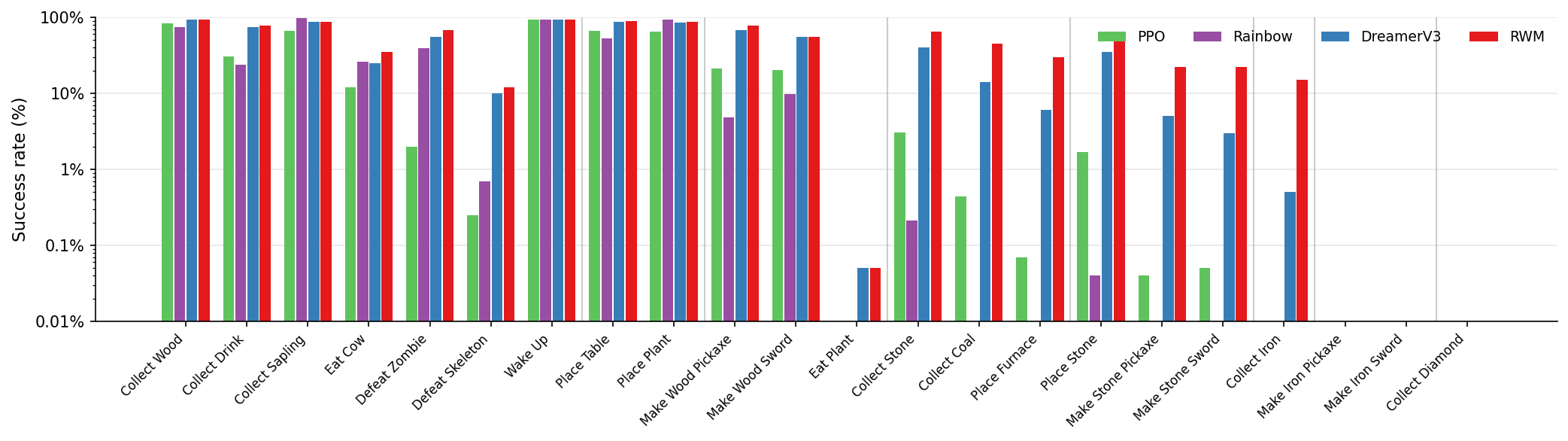}
    \caption{Agent ability spectrum for Crafter (ordered left to right by number of prerequisites for an achievement). The y-axis is shown in log scale. RWM achieves comparable performance on common early-stage achievements and stronger performance on harder long-horizon achievements involving resource collection and tool crafting.}
    \label{fig:crafter_achievements}
\end{figure*}

\textbf{Per-achievement Results.} Figure~\ref{fig:crafter_achievements} shows that RWM is similar to DreamerV3 on early achievements with few prerequisites (wood, drinking, sapling, waking up, placing a table), while higher on achievements with longer dependency chains (coal, furnace, stone tools, iron), which is consistent with the Minecraft results.

\section{Implementation Details on the Web}
\label{app:web_protocol}
\label{app:graph_web}

\subsection{Web embodied graph.} We follow the official observation and action spaces~\cite{liu2025visualagentbench,zhou2024webarena}. Candidate nodes are the DOM elements of the current page, with their tag, ARIA role, input type, text, value and instruction strings, normalized bounding box, and corresponding flags. A virtual page node replaces the agent node. Edges connect parent and child elements in both directions and link native labels to the controls they activate. An interaction $A=(i,v_{tar})$ is one of click, fill, press, scroll, or select applied to a target node, together with its payload (typed text, key, option, or scroll amount).

\subsection{Construction and Learning}
TA-EGC scores the DOM candidates with the same task, novelty and coverage heads as in embodied control and admits the Top-$P$ into the graph. Instead of spatial proximity, a candidate is matched to an existing node by its document, element reference and tag. The task label marks the targets of the next $W{=}16$ recorded interactions. Novelty and coverage use the rarity of the node type and the rarity of its position on the screen, respectively. All labels are weighted uniformly, since the Web policy is fixed and provides no advantage. The prediction heads decode the state flags, input values and bounding boxes of the elements, and TA-EGC and EG-RSSM are trained jointly.

\subsection{Navigation Head}
On the Web, the structural outcome in $\mathcal{L}_{topo}$ is the page that an interaction leads to: a click either stays on the current page, where changes of elements are captured by the predicted flags, or loads a new page. We predict this outcome with a navigation head, a softmax classifier on the frozen EG-RSSM latents. Its classes are \textsc{stay} or the site routes of the pages seen in training.

\subsection{Decision procedure.} At each step, WebRL proposes 3 candidate actions from the official prompt. The world model forecasts each candidate from the shared posterior. The verifier (Qwen2.5-VL-32B-Instruct) receives the task, the current screenshot, the observed action history, the candidates and their forecasts, and returns one candidate. Unparsable replies fall back to WebRL's top candidate. The verifier is greedily decoded with at most 768 output tokens. All three variants use the same procedure, prompts, and budgets. For the variant without forecasts, the candidates are shown without predicted consequences.

\subsection{Environment and scoring.} Each episode starts from a newly launched CMS container with a fresh login. The tasks follow the official configuration and the early-stop rules, and the browser window is set to $1280\times720$ pixels. Judge-graded tasks are scored with the official judge prompt. Episodes ending with an environment error count as failures.

\end{document}